\documentclass[letterpaper]{article} 
\usepackage{aaai23}  
\usepackage{times}  
\usepackage{helvet}  
\usepackage{courier}  
\usepackage[hyphens]{url}  
\usepackage{graphicx} 
\usepackage{booktabs}
\usepackage{multirow}
\usepackage{natbib}  
\usepackage{caption} 
\usepackage{algorithm}
\usepackage{algorithmic}
\usepackage{subcaption}

\usepackage{amsmath,amssymb}
\usepackage{adjustbox}
\usepackage{newfloat}
\usepackage{listings}

\usepackage{makecell}

\DeclareCaptionStyle{ruled}{labelfont=normalfont,labelsep=colon,strut=off} 
\floatstyle{ruled}
\newfloat{listing}{tb}{lst}{}
\floatname{listing}{Listing}
\usepackage{colortbl}

\usepackage{xspace}

\title{When hard negative sampling meets supervised contrastive learning}

\author{
    Zijun Long\textsuperscript{\rm 1},
    George Killick \textsuperscript{\rm1},
    Richard McCreadie\textsuperscript{\rm 1},
    Gerardo Aragon Camarasa\textsuperscript{\rm 1},
    Zaiqiao Meng\textsuperscript{\rm 1},
}
\affiliations{
    \textsuperscript{\rm 1}University of Glasgow \\
    \{zijun.long, george.killick, richard.mccreadie, gerardo.aragoncamarasa, zaiqiao.meng\}@glasgow.ac.uk\\
    
}

\usepackage{bibentry}

\begin{document}

\maketitle

\begin{abstract}
State-of-the-art image models predominantly follow a two-stage strategy: pre-training on large datasets and fine-tuning with cross-entropy loss. Many studies have shown that using cross-entropy can result in sub-optimal generalisation and stability. While the supervised contrastive loss addresses some limitations of cross-entropy loss by focusing on intra-class similarities and inter-class differences, it neglects the importance of hard negative mining. We propose that models will benefit from performance improvement by weighting negative samples based on their dissimilarity to positive counterparts. In this paper, we introduce a new supervised contrastive learning objective, SCHaNe, which incorporates hard negative sampling during the fine-tuning phase. Without requiring specialized architectures, additional data, or extra computational resources, experimental results indicate that SCHaNe outperforms the strong baseline BEiT-3 in Top-1 accuracy across various benchmarks, with significant gains of up to $3.32\%$ in few-shot learning settings and $3.41\%$ in full dataset fine-tuning. Importantly, our proposed objective sets a new state-of-the-art for base models on ImageNet-1k, achieving an 86.14\% accuracy. Furthermore, we demonstrate that the proposed objective yields better embeddings and explains the improved effectiveness observed in our experiments.
\end{abstract}

\section{Introduction}
\label{sec:intro}

Achieving state-of-the-art performance in image classification tasks often employs models that are initially pre-trained on auxiliary tasks and then fine-tuned with cross-entropy loss~\cite{RN83,beit3,RN82}. This prevalent approach, however, leads to sub-optimal results due to the limitations of cross-entropy. Specifically, the measure of KL-divergence between one-hot label vectors and model outputs can hinder generalization~\cite{DBLP:conf/icml/LiuWYY16, DBLP:journals/corr/abs-1906-07413} and show sensitivity to noisy labels~\cite{DBLP:journals/corr/abs-1901-08360,DBLP:conf/icml/LiuWYY16} or adversarial samples~\cite{RN97,RN98}. Various techniques have emerged to address these problems, such as knowledge distillation~\cite{DBLP:journals/corr/HintonVD15}, self-training~\cite{DBLP:journals/corr/abs-1905-00546}, Mixup~\cite{DBLP:journals/corr/abs-1710-09412}, CutMix~\cite{DBLP:journals/corr/abs-1905-04899}, and label smoothing~\cite{DBLP:journals/corr/SzegedyVISW15}. However, in scenarios like few-shot learning, cross-entropy loss can still limit the performance of models. Despite the proposed solutions, such as extended fine-tuning epochs and specialized optimizers~\cite{DBLP:journals/corr/abs-2006-05987, DBLP:journals/corr/abs-2006-04884}, which mitigate the limitations of cross-entropy to some extent, they introduce new challenges, such as extended training durations, and the increased complexity persists.

Amidst these challenges, contrastive learning emerges as a potential solution. Recently, it has shown promise in image model training, such as SimCLR~\cite{RN89} showcasing strong performance, especially in few-shot learning. Building upon such foundations, there is growing interest in combining task labels with contrastive learning, seeking to improve vanilla cross-entropy pre-training approaches~\cite{RN91}. Motivated by these successes, we ask: \textit{Can contrastive learning prove beneficial during the fine-tuning phase, especially in addressing the shortcomings of cross-entropy loss?}

To fully exploit the potential of contrastive learning, it is crucial to adhere to its foundational principles: promoting similarity among positive pairs (intra-class data points) and maximizing differences for negative pairs (inter-class data points). Task-specific label information becomes indispensable in correctly identifying these positive pairs. A critical observation, however, is that many current supervised contrastive learning methods~\cite{RN91,suconlan} and unsupervised approaches~\cite{RN89,npairloss,DBLP:journals/corr/abs-1810-06951} often overlook the nuanced selection of negative samples, typically treating them in a uniform manner. This generalization misses out on the demonstrated benefits of leveraging ``hard" negative samples for accelerated corrective learning, as underscored by several studies~\cite{DBLP:journals/corr/SongXJS15,DBLP:journals/corr/SchroffKP15}. In response to this evident gap in the research, we present a new fine-tuning objective. Our approach integrates a label-aware contrastive learning mechanism with hard negative sampling, optimizing the selection of both positive and negative samples to achieve superior model performance.

\begin{figure*}
    \centering
    \includegraphics[width=10cm]{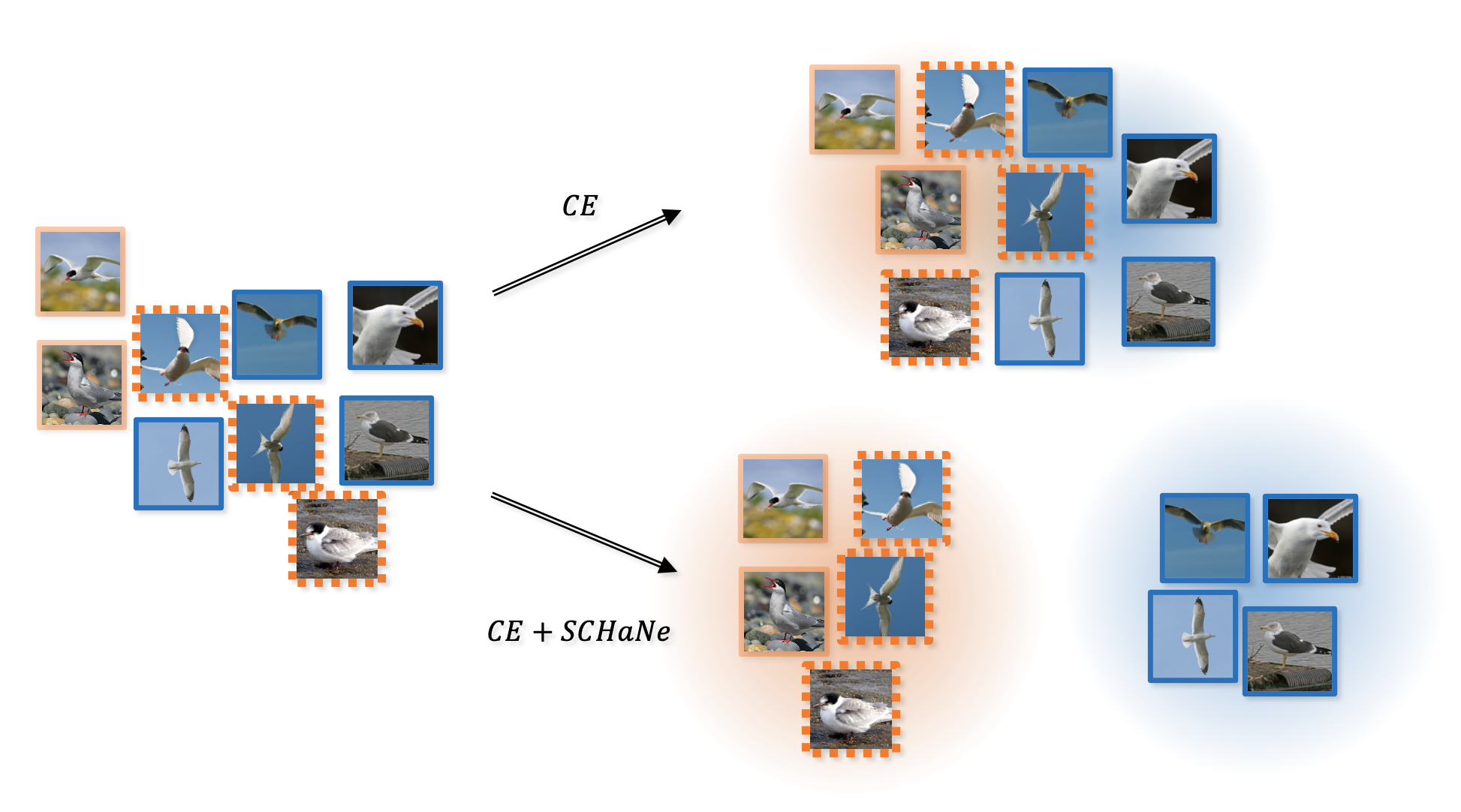}
    \caption{Our proposed loss objective integrates a Supervised Contrastive learning with Hard Negative sampling term, named SCHaNe, and a Cross-Entropy (CE) term. It is formulated to pull examples from the same class closer, while pushing those from different classes apart. As illustrated using samples from the CUB-200-2011: on the left, embeddings of positive examples are outlined in blue, while negative examples are highlighted in orange. Negatives outlined with thick dashed borders are identified as hard negatives by our proposed objective, receiving greater emphasis compared to other negatives. This underscores their marked visual similarity to the positive counterparts. On the right, the embeddings of two classes after fine-tuning with different objectives are shown. Through our SCHaNe loss function, we generate discernible separations in latent embeddings between positive and negative instances. This is achieved by emphasizing the hard negatives, a feature not present in the traditional CE loss.}
    \label{fig:embeddingvis}
\end{figure*}

Through empirical validation, as highlighted in Figure~\ref{fig:embeddingvis}, we demonstrate that our proposed objective allows the encoder to generate more distinct embeddings by focusing on hard negative samples, leading to statistically significant improvements in performance. Our main contributions are:

\begin{itemize}
    \item \textbf{Introduction of a Novel Objective}: We propose the SCHaNe contrastive loss function that results in distinct embeddings by prioritizing hard negative samples. This objective results in enhanced model performance without the need for specialized architectures or additional resources.
    \item \textbf{Superior Performance in Few-Shot Learning}: When benchmarked against the very strong BEiT-3 baseline, our approach demonstrates a significant improvement in Top-1 accuracy across diverse datasets. This improvement is especially evident in challenging 1-shot learning scenarios. Specifically, we observe a boost of up to 3.32\% in Top-1 accuracy on the CIFAR-FS~\cite{DBLP:conf/iclr/cifarfs} dataset compared to the BEiT-3 baseline. 
    \item \textbf{State-of-the-Art in Full Dataset Fine-Tuning}: Our method consistently outperforms competitors across various image datasets, establishing a new state-of-the-art for base models on ImageNet-1k. Furthermore, our approach achieves a notable increase of 3.41 \% on the iNaturalist 2017 dataset~\cite{DBLP:conf/cvpr/inat2017} compared to the baseline BEiT-3.
    \item \textbf{Pioneering Integration}: To the best of our knowledge, ours is the first work that combines explicit hard negative sampling with supervised contrastive objectives for refining pre-trained image models.
    \item \textbf{Detailed Analysis of Effect of the Proposed Contrastive Loss Function}: A detailed analysis (Section~\ref{sec:analysis}) reveals that BEiT-3, when refined with our proposed objective, produces enhanced representations, further elucidating the improved efficacy observed in our experiments.
\end{itemize}

\section{Related work}
\subsection{Limitations of the Cross-Entropy Loss}
Cross-entropy loss has long been the default setting for many deep neural models due to its ability to optimize classification tasks effectively. However, recent research has revealed several inherent drawbacks~\cite{ DBLP:conf/icml/LiuWYY16, DBLP:journals/corr/abs-1906-07413}. Specifically, models trained with cross-entropy tend to exhibit poor generalization capabilities. This vulnerability comes from insufficient margins, making the model more susceptible to noisy labels \cite{DBLP:journals/corr/abs-1901-08360,DBLP:conf/icml/LiuWYY16} and adversarial examples \cite{RN97,RN98}. These deficiencies underscore the need for alternative loss functions that offer better robustness and discrimination capabilities.

\subsection{Contrastive Learning: Overview}
Contrastive learning, originally introduced by \cite{becker1992self}, aims to improve the distinction between samples by minimizing the distance between positive pairs (instances of the same class) and maximizing the distance for negative pairs (instances of different classes). This approach has been transformative, especially in the realm of computer vision. For example, SimCLR \cite{RN89} leveraged instance-level comparisons for image classification in an unsupervised manner, setting performance benchmarks on datasets such as ImageNet ILSVRC-2012 \cite{RN91}. Furthermore, the adoption of label information, as explored by \cite{RN81}, propelled contrastive learning to a new level, revealing it as a viable alternative, if not superior, to cross-entropy loss. This advancement highlighted the stability of supervised contrastive loss across various hyperparameters and data augmentations.

\subsection{The Criticality of Negative Mining}
While the benefits of positive sample identification are well-established, the nuanced role of negative samples, particularly hard negatives, has been comparatively underexplored. Some pioneering works, such as \cite{DBLP:journals/corr/abs-2007-00224}, tackled the challenge of distinguishing true negatives from a sea of potential negatives, invoking unlabelled positive learning principles to approximate the true negative distribution. The integration of mixup techniques~\cite{DBLP:journals/corr/abs-1710-09412} with contrastive learning, as proposed by~\cite{DBLP:journals/corr/abs-2010-01028}, further emphasized the potential of mining hard negatives in latent space. Numerous studies \cite{DBLP:journals/corr/SongXJS15,DBLP:journals/corr/KumarHC0D17,DBLP:journals/corr/WuMSK17,DBLP:journals/corr/abs-1810-06951,suh2019stochastic} have consistently underlined the pivotal role of these hard negatives in refining the discriminative power of embeddings.

Drawing from these advances, our work seeks to amalgamate the strengths of supervised contrastive learning with the discernment of hard negative sampling. We propose an innovative loss function that not only recognizes the importance of positive samples, but also accentuates the criticality of effectively weighting negative samples, particularly the hard negatives, for improved model performance.

\section{Approach}\label{sec:Method}

We propose a novel objective for fine-tuning pre-trained image models that integrates a Supervised Contrastive loss with Hard Negative sampling (SCHaNe) and Cross-Entropy loss (CE). This loss aims to encapsulate similarities between instances of the same class (positive samples) using label information and contrasts them against instances from other classes (negative samples), with a particular emphasis on hard negative samples. The foundation of this loss draws from the premise that the training efficacy of negative samples varies between soft and hard samples. We argue that weighting negative samples based on their dissimilarity to positive samples is more effective. This allows the model to prioritize distinguishing between positive samples and those negative samples that the embedding deems similar to the positive ones, ultimately enhancing the overall performance.

\subsection{Proposed Loss Functions}

The overall loss function in our method is a weighted combination of the vanilla cross-entropy loss and our novel SCHaNe loss, as expressed in Equation~\ref{eq:weighted}:

\begin{equation}
    \mathcal{L} = (1-\lambda )\mathcal{L}_\mathrm{CE} + \lambda \mathcal{L}_\mathrm{SCHaNe}
    \label{eq:weighted}
\end{equation}

In Equation~\ref{eq:weighted}, the term \(\mathcal{L}_\mathrm{CE}\) represents the cross-entropy loss, while \(\mathcal{L}_\mathrm{SCHaNe}\) symbolizes our proposed supervised contrastive loss. $\lambda$ represents a scalar weighting hyperparameter tuned for each specific task and setting. $\lambda$ determines the relative importance of each of the two losses. If $\lambda$ is 1, only SCHaNe loss is considered, if it is 0, only cross-entropy loss is considered. Any value in between gives a mixture of the two.

To provide context for \(\mathcal{L}_\mathrm{CE}\), we refer to the vanilla definition of the multi-class cross-entropy loss, detailed in Equation~\ref{eq:ce}:

\begin{equation}
\mathcal{L}_\mathrm{CE} = -\frac{1}{N} \sum_{i=1}^{N} \sum_{c=1}^{C} z_{i,c} \log(\hat{z}_{i,c})
\label{eq:ce}
\end{equation}

In Equation~\ref{eq:ce}, \(z_{i,c}\) and \(\hat{z}_{i,c}\) represent the label and the model's output probability for the \(i\)th instance belonging to class \(c\), respectively.

We then introduce \(\beta\), a weight factor for each negative sample, as described in Equation~\ref{reweight}:

\begin{equation}
\beta =\exp(z_{i}\cdot z_{k}/\tau) \displaystyle\frac{\lvert \mathcal{D}^{*\textbf{-}}_{-z_i}\lvert}{\displaystyle\sum\limits_{z_{k}\in \mathcal{D}^{*\textbf{-}}_{-z_i}} {\exp(z_{i}\cdot z_{k}/\tau)}}
\label{reweight}
\end{equation}

The weight is based on the dot product (or similarity) of the embeddings and is normalized by a temperature parameter \(\tau\). The purpose of this equation is to give more emphasis to the hard negative samples, i.e., the negatives that the current embeddings are the most similar to the positive samples.

Following the description of \(\beta\), it is appropriate to introduce the formal definition of our SCHaNe loss, illustrated in Equation~\ref{eq:contrastive_loss_fn}:

\begin{equation}
\resizebox{0.4\textwidth}{!}{$
\mathcal{L}_{\text{SCHaNe}} = \displaystyle\sum\limits_{z_{i}\in \mathcal{D}^{*}} \log \frac{-1}{\lvert \mathcal{D}^{*\textbf{+}}_{-z_i}\lvert}  \frac{\displaystyle\sum\limits_{z_{p}\in \mathcal{D}^{*\textbf{+}}_{-z_i}} \exp(z_{i}\cdot z_{p}/\tau)}{\displaystyle\sum\limits_{z_{p}\in \mathcal{D}^{*\textbf{+}}_{-z_i}} \exp(z_{i}\cdot z_{p}/\tau) + \displaystyle\sum\limits_{z_{k}\in \mathcal{D}^{*\textbf{-}}_{-z_i}} \beta \exp(z_{i}\cdot z_{k}/\tau)}
$}
\label{eq:contrastive_loss_fn} 
\end{equation}


In Equation~\ref{reweight} and \ref{eq:contrastive_loss_fn}, $\mathcal{D}^{*}$ represents the entire mini-batch composed of an embedding $z$ for each image view (or anchor) $i$. Therefore, $z_i\in\mathcal{D}^{*}$ is a set of embeddings within the mini-batch. The superscripts $+$ and $-$, e.g. $\mathcal{D}^{*\textbf{+}}$, denote sets of embeddings consisting only of positive and negative examples, respectively, for the current anchor within the mini-batch. The term $\lvert \mathcal{D}^{*\textbf{+}}_{-z_i}\lvert$ represents the cardinality of the positive set for the current anchor, while the subscript $-z_i$ denotes that this set excludes the embedding $z_i$. The symbol $\cdot$ represents the dot product. $\tau$ is a scalar temperature parameter controlling class separation. A lower value for $\tau$ encourages the model to differentiate positive and negative instances more distinctly.

Equation~\ref{eq:contrastive_loss_fn} is the core of our contribution. It tries to minimize the distance between embeddings of positive pairs (the same class) and maximize the distance between the anchor and negative samples, especially the hard negatives. This equation has two main parts: (1) The numerator focuses on the positive samples and aims to make their embeddings close to the anchor embedding. (2) The denominator contains both positive samples and weighted negative samples. The goal here is to ensure the anchor's embedding is distant from negative samples, especially the hard ones. The weights (or importance) of these negative samples are given by the previously calculated \(\beta\).

\subsection{Contrastive Loss Function Analysis}
To enhance the effectiveness of contrastive learning, our approach is rooted in three foundational principles that have been empirically supported by numerous studies~\cite{RN89,RN91,DBLP:journals/corr/abs-2007-00224,DBLP:journals/corr/abs-2010-01028}: 
\begin{itemize}
 \item \textbf{True Positives}: Samples must strictly share the same label as the anchor \(x\), which drives the model to discern inherent similarities.
\item \textbf{True Negatives}: Samples with distinct labels from the anchor \(x\) ensure computational focus on genuinely contrasting pairs.
\item \textbf{Emphasis on Hard Negatives}: The crux of effective contrastive learning is the model's ability to discern between closely related samples. Therefore, negative samples that the model's current embedding perceives as akin to the anchor, termed as "hard" negative samples, are the most instructive. They push the model's boundaries, facilitating a more refined feature extraction.
\end{itemize}

Reflecting on these principles, our proposed supervised contrastive loss with hard negative sampling achieves:
\begin{itemize} \item \textbf{Robust Positive/Negative Differentiation}: By leveraging explicit label information, as encapsulated in equation~\ref{eq:contrastive_loss_fn}, we ensure a categorical distinction between true positive and true negative samples. This not only prevents the model from being misled by incorrectly labelled samples but also reinforces the core philosophy of contrastive learning. The objective is dual: to reduce the distance between embeddings of positive pairs and to widen the gap for negative pairs, ensuring robust class separation.
\item \textbf{Discriminating Fine Detail with Hard Negatives}: Our methodology adjusts the weighting of negative samples based on their similarities to positive instances, as defined in equation~\ref{reweight}. This nuanced approach ensures that the model does not merely differentiate between glaringly distinct samples but also hones its skills on the more challenging, closely related negative samples. Such an approach paves the way for a model that is not just accurate, but also discerning in real-world scenarios where differences between classes might be minimal. 
\end{itemize}

This refined approach to contrastive learning promises to introduce models with enhanced discriminative power, fine-tuned to the intricate details of the data on which they are trained.

\subsection{Relationship to Other Contrastive Learning Approaches}

Contrastive learning, especially self-supervised variants, has recently surged in popularity for learning robust representations, especially in the computer vision domain~\cite{RN89,DBLP:journals/corr/abs-1911-05722,DBLP:conf/eccv/TianKI20,DBLP:conf/cvpr/KolesnikovZB19}. Such approaches, which function without labelled data, are particularly appealing for pre-training given the vast amount of available unlabelled data. Notable methods such as SimCLR~\cite{RN89} achieve competitive performance even without labels, although fully outperforming supervised methods remains challenging. A few studies have considered hard negative sampling in an unsupervised manner. For instance, \cite{DBLP:journals/corr/abs-2007-00224} addresses sampling bias with a debiased loss, while another work~\cite{ DBLP:conf/iclr/RobinsonCSJ21} leverages intricate approximations to determine the distribution of negative samples. However, these complicated methods introduce unnecessary complexity, either in computation or model architecture, and their results still do not represent the true negative distribution due to the absence of labels. 

This unsupervised manner raises the question on effectiveness of these contrastive learning frameworks, especially during fine-tuning. With many downstream datasets having labels, ignoring them can be damaging to performance. For instance, \cite{RN91} proposed a supervised contrastive loss that exploits label information during pre-training. Still, it relies on implicit hard negative mining, which exhibits constraints. In the realm of natural language processing, \cite{suconlan} effectively applied supervised contrastive loss during fine-tuning but also did not leverage label data for hard negative mining. This leaves a research gap: exploring the effectiveness of supervised contrastive learning and hard negative sampling, especially in the context of fine-tuning in the vision domain.

\subsection{Representation Learning Framework}

Our framework is built on three main components:

\begin{itemize}
    \item \textbf{Data Augmentation module,} ${Aug(\cdot)}$: This component creates two different views of each sample $x$, called $\tilde{x}={Aug(x)}$. This means that every sample will have at least one similar sample (positive pair) in a batch during fine-tuning.
    
    \item \textbf{Encoder Network,} ${Enc(\cdot)}$: This network encodes the input data, $x$, into a representation vector, $r=Enc(x)$. Each of the two different views of the data is fed into the encoder separately. 
    
    \item \textbf{Classification head,} $Head(\cdot)$: This maps the representation vector, $r$, to a size that matches the number of classes in the target task. Primarily a linear layer, and we use its output to calculate the cross-entropy loss.
\end{itemize}

Our SCHaNe loss can be applied using a wide range of encoders, such as BEiT-3~\cite{beit3} for computer vision applications and models such as BERT~\cite{RN53} for natural language processing tasks. In this paper, our primary focus is on fine-tuning of pre-trained image models. We validate our approach using both few-shot and full dataset experimental settings, aiming to compare the effectiveness of our proposed objective with the vanilla cross-entropy loss. Following the method in \cite{RN89}, every image in a batch is altered to produce two separate views (anchors). Views that have the same label as the anchor are considered positive, while the rest are viewed as negative. The encoder output, represented by $y_{i} = {Enc(z_{i})}$, is used to calculate the contrastive loss. In contrast, the output from the classification head, denoted as $p_{i} = {Head(Enc(z_{i}))}$, is used for the cross-entropy loss. Building on insights from prior studies, we have incorporated L2 normalization on encoder outputs, a strategy demonstrated to significantly enhance performance~\cite{DBLP:conf/eccv/TianWKTI20}.

\begin{table*}[]
\centering
\resizebox{1\textwidth}{!}{\begin{tabular}{llccccccccc}

\toprule
            &                           & \multicolumn{2}{c}{CIFAR-FS}                                    & \multicolumn{2}{c}{FC100}                                       & \multicolumn{2}{c}{miniImageNet}                                & \multicolumn{2}{c}{tieredImageNet}                              \\ \hline
Backbone    & Method                                 & 1-shot                         & 5-shot                         & 1-shot                         & 5-shot                         & 1-shot                         & 5-shot                         & 1-shot                         & 5-shot                         \\ \hline
64-64-64-64 & Prototypical Networks \cite{DBLP:conf/nips/SnellSZ17}                       & 55.50$\pm$0.70                     & 72.00$\pm$0.60                     & 35.30$\pm$0.60                     & 48.60$\pm$0.60                     & 49.42$\pm$0.78                     & 68.20$\pm$0.66                     & 53.31$\pm$0.89                     & 72.69$\pm$0.74                     \\
ResNet-12   & MetaOptNet \cite{DBLP:conf/cvpr/LeeMRS19}                                   & 72.60$\pm$0.70                     & 84.30$\pm$0.50                     & 41.10$\pm$0.60                     & 55.50$\pm$0.60                     & 62.64$\pm$0.61                     & 78.63$\pm$0.46                     & 65.99$\pm$0.72                     & 81.56$\pm$0.53                     \\
WRN-28-10   & CE \cite{DBLP:conf/iclr/DhillonCRS20}                                          & \multicolumn{1}{l}{68.72$\pm$0.67} & \multicolumn{1}{l}{86.11$\pm$0.47} & \multicolumn{1}{l}{38.25$\pm$0.52} & \multicolumn{1}{l}{57.19$\pm$0.57} & \multicolumn{1}{l}{57.73$\pm$0.62} & \multicolumn{1}{l}{78.17$\pm$0.49} & \multicolumn{1}{l}{66.58$\pm$0.70} & \multicolumn{1}{l}{85.55$\pm$0.48} \\
WRN-28-10   & Transductive fine-tuning \cite{DBLP:conf/iclr/DhillonCRS20}                     & 76.58$\pm$0.68                     & 85.79$\pm$0.50                     & 43.16$\pm$0.59                     & 57.57$\pm$0.55                     & 65.73$\pm$0.68                     & 78.40$\pm$0.52                     & 73.34$\pm$0.71                     & 85.50$\pm$0.50                     \\
ResNet-12   & Distillation \cite{DBLP:conf/eccv/TianWKTI20}                                & 73.90$\pm$0.80                     & 86.90$\pm$0.50                     & 44.60$\pm$0.70                     & 60.90$\pm$0.60                     & 64.82$\pm$0.60                     & 82.14$\pm$0.43                     & 71.52$\pm$0.69                     & 86.03$\pm$0.49                     \\ \hline

BEIT3         & CE                                       & 83.68$\pm$0.80     & 93.01$\pm$0.38    & 66.35$\pm$0.95   & 84.33$\pm$0.54   & 90.62$\pm$0.60       & 95.77$\pm$0.28      & 84.84$\pm$0.70        & 94.81$\pm$0.34       \\
BEIT3         & CE + SCHaNe                        & \textbf{87.00$\pm$0.70}     & \textbf{93.77$\pm$0.36}    & \textbf{69.87$\pm$0.91}   & \textbf{87.06$\pm$0.52}   & \textbf{92.35$\pm$0.53 }      & \textbf{96.78$\pm$0.23}      & \textbf{87.24$\pm$0.62}        & \textbf{96.09$\pm$0.29}      \\ \Xhline{2\arrayrulewidth}
\end{tabular}}
\caption{\textbf{Comparison to baselines on the few-shot learning setting.} Average few-shot classification accuracies(\%) with 95\% confidence intervals on test splits of four few-shot learning datasets. Backbones in the form of a-b-c-d refer to the number of filters in each layer of a CNN.}
\label{tab:FS}
\end{table*}

\section{Experimental Setup}
To evaluate our proposed objective, we benchmark its performance across two primary image classification scenarios: few-shot learning and full dataset fine-tuning.

\looseness -1 \textbf{Few-shot Learning Setting:} We employ four prominent few-shot image classification benchmarks for our evaluations: CIFAR-FS~\cite{DBLP:conf/iclr/cifarfs}, FC100~\cite{DBLP:conf/nips/fc100}, miniImageNet~\cite{DBLP:conf/nips/miniimagenet}, and tieredImageNet~\cite{DBLP:conf/iclr/tieredimagenet}. The CIFAR-FS and FC100 datasets originate from CIFAR~\cite{krizhevsky2009learning}, while the miniImageNet and tieredImageNet datasets are derivatives of Imagenet-1k~\cite{DBLP:conf/cvpr/imagenet}. We adhere to the widely used splitting protocol proposed in~\cite{DBLP:conf/iclr/cifarfs,DBLP:conf/nips/fc100, DBLP:conf/iclr/RaviL17} to ensure fair comparisons with baselines. In these experiments, we evaluate our methodology on 3 runs and report both the median Top-1 accuracy and the 95\% confidence interval of this median on the test set. Each run consists of 3,000 randomly sampled tasks from the test set, maintaining a consistent query shot count of 15 for all tests.

\looseness -1 \textbf{Full Dataset Fine-tuning Setting:} We leverage eight renowned, publicly available image classification datasets for our full dataset fine-tuning evaluations: CIFAR-100~\cite{krizhevsky2009learning}, CUB-200-2011~\cite{WahCUB_200_2011}, Caltech-256~\cite{griffin2007caltech}, Oxford 102 Flower~\cite{DBLP:conf/icvgip/flowers}, Oxford-IIIT Pet~\cite{DBLP:conf/cvpr/pets}, iNaturalist 2017~\cite{DBLP:conf/cvpr/inat2017}, Places365~\cite{DBLP:conf/nips/places}, and ImageNet-1k~\cite{DBLP:conf/cvpr/imagenet}. Following established conventions, we maintain train/test splits in line with prior works~\cite{krizhevsky2009learning,WahCUB_200_2011,griffin2007caltech,DBLP:conf/icvgip/flowers,DBLP:conf/cvpr/pets,DBLP:conf/cvpr/inat2017,DBLP:conf/nips/places,DBLP:conf/cvpr/imagenet} to ensure fair comparisons. The performance metrics for these experiments denote the mean Top-1 accuracy across three distinct runs in the test set, each initialized using different random seeds. 

\textbf{Implementation Details:} We conduct training over 100 epochs for the few-shot datasets and 50 epochs for the full datasets. We utilize the base model of the state-of-the-art BEiT-3 as our encoder backbone, because of its state-of-the-art performance on ImageNet-1k. For data augmentation, we employ AutoAugment~\cite{DBLP:conf/cvpr/CubukZMVL19}, which has proven to be highly effective for supervised contrastive learning~\cite{RN91}. Throughout all fine-tuning experiments, we choose the Adam optimizer, with a learning rate set at $1\times10^{-4}$, a weight decay of $0.05$, and a batch size of $1024$ (unless specified otherwise). We also integrate a dropout rate of 0.1. In experiments that incorporate the SCHaNe term, we perform an extensive grid-based hyperparameter search on the validation set, adjusting $\lambda$ across the range \{0, 0.1, 0.3, 0.5, 0.7, 0.9, 1.0\} and $\tau$ within \{0.1, 0.3, 0.5, 0.7\}. Our observations predominantly support the use of the hyperparameter combination $\tau=0.5$ and $\lambda=0.9$ in all evaluated scenarios, as these configurations persistently achieved the highest validation accuracies.

\begin{table*}[]
\centering
\resizebox{1.0\textwidth}{!}{\begin{tabular}{@{}llcccccccc@{}}
\Xhline{2\arrayrulewidth}
Model & Fine-tune method  & CIFAR-100      & CUB-200        & caltech256     & Oxford Flowers & Pet            & iNat2017       & Places365 & ImageNet-1k    \\ \midrule
ViT-B & CE                                       & 87.13          & 76.93          & 90.92          & 90.86          & 93.81          & 65.26          & 54.06     & 77.91          \\
MAE   & CE                                        & 87.67          & 78.46          & 91.82          & 91.67          & 94.05          & 70.50          & 57.90      & 83.60          \\
BEiT-3 & CE                                       & 92.96          & 98.00          & 98.53          & 94.94          & 94.49          & 72.31          & 59.81         & 85.40          \\
BEiT-3 & CE + SCHaNe                        & \textbf{93.56} & \textbf{98.93} & \textbf{99.41} & \textbf{95.43} & \textbf{95.62} & \textbf{75.72} & \textbf{62.22}         & \textbf{86.14} \\ \Xhline{2\arrayrulewidth} 
\end{tabular}}
\caption{\textbf{Comparison to baselines on full dataset fine-tuning setting.} Classification accuracies on eight various datasets.}
\label{tab:FD}
\end{table*} 

\section{Experimental results}

\textbf{Few-shot Results:}
The few-shot learning results are shown in Table \ref{tab:FS}. This evaluation emphasizes the model's ability to generalize well with limited labelled data for each class. Our BEiT-3-CE baseline (fine-tuned with vanilla cross-entropy) already outperforms all other baselines by at least 9.78\% in Top-1 accuracy. The proposed loss function brings a further consistent and significant improvement to BEiT-3 (denoted BEiT-3-SCHaNe) in Top-1 accuracy across the four selected few-shot datasets over the BEiT-3-CE baseline. In particular, in a 1-shot learning setting, the largest improvement is observed on FC100 when compared to the BEiT-3 baseline; a boost of $3.32\%$ in accuracy from $66.35\%$ to $69.87\%$. On average, there is an increase of $2.7\%$ across all datasets. For the 5-shot learning setting, the improvements are again notable. We observe the largest increase from $84.33\%$ to $87.06\%$ on FC100, and on average, an increase of $1.4\%$ is observed in the datasets. The performance difference between 1-shot learning and 5-shot learning indicates that our proposed objective is more effective as the number of positive samples for each class decreases. It is worth noting that there is a statistically significant improvement when comparing the results of BEiT-3-SCHaNe with BEiT-3-CE across four few-shot datasets when using the paired t-test with 
$p<0.01$. Additionally, there is reduced variability in the results, as evidenced by tighter confidence intervals for Top-1 accuracy compared to BEiT-3-CE. Overall, these outcomes underline the efficacy of integrating our proposed contrastive loss with hard negative sampling (SCHaNe) into the vanilla cross-entropy function, refining BEiT-3's pre-trained representations for few-shot learning tasks.

\textbf{Full Dataset Results:} 
Table \ref{tab:FD} presents the results of full dataset fine-tuning, which offers further evidence of the applicability of the proposed objective beyond few-shot scenarios. Our proposed objective consistently boosts classification accuracy across varied image classification datasets when compared to three state-of-the-art image models with vanilla cross-entropy fine-tuning. In some instances, the increase in performance by our proposed objective (BEiT-3-SCHaNe), is modest, given the particularly strong baseline already established by BEiT-3-CE, as seen with a modest rise from $98.00\%$ to $98.93\%$ on CUB-200-2011. However, for more challenging datasets, such as iNaturalist2017, we observe a significant increase in classification accuracy, rising from $72.31\%$ to $75.72\%$. We believe that the difference in the increase in performance is due to the dataset's extensive class variety, e.g., 5087 classes in iNaturalist2017 vs. 200 classes in CUB-200-2011. A similar increase is evident in ImageNet-1k, where the accuracy increased from $85.40\%$ to $86.14\%$. It is important to highlight that the $85.40\%$ accuracy was the highest for a base model on the leaderboard at the time of writing this paper~\footnote{https://paperswithcode.com/sota/image-classification-on-imagenet}, implying that $86.14\%$ sets a new state-of-the-art record. Furthermore, we observe a statistically significant improvement when comparing the BEiT-3-SCHaNe results with BEiT-3-CE on the tested datasets when using the paired t-test with $p<0.05$. In essence, our results emphasize the prowess of the SCHaNe loss in amplifying the performance of pre-trained models during subsequent fine-tuning across a range of datasets. Importantly, our approach is intuitive and achieves these results without the need for specialized architectures, extra data, or increased computational overhead, making it a compelling and powerful alternative to the vanilla cross-entropy loss.

\section{Analysis}\label{sec:analysis}

\subsection{Optimizing $\lambda$: Bridging Cross-Entropy and Contrastive Learning}
\begin{figure}
     \centering
     \includegraphics[width=1\linewidth]{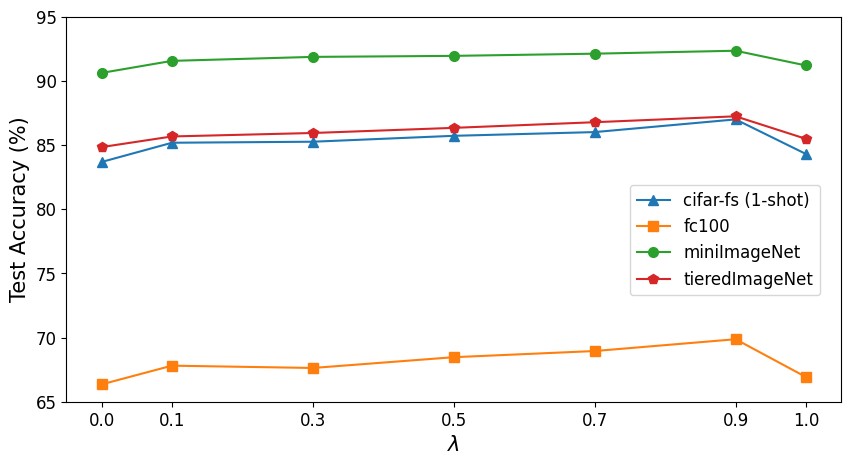}
     \caption{\textbf{Evaluation of the impact of the $\lambda$ hyperparameter.} One-shot results on CIFAR-FS with $\lambda$ values ranging from {0, 0.1, 0.3, 0.5, 0.7, 0.9, 1.0}.}
     \label{fig:lambdasweep}
 \end{figure}

Our proposed loss function incorporates a hyperparameter, $\lambda$, to balance the contributions of the cross-entropy term and the proposed SCHaNe terms, shown in Equation ~\ref{eq:weighted}. Specifically, for $\lambda = 0$, the loss is equivalent to the vanilla cross-entropy, while for $\lambda = 1$, the loss is exclusively the SCHaNe term.
To understand the influence of $\lambda$, we evaluated its effect on classification accuracy in a one-shot setting on four few-shot learning datasets. Figure \ref{fig:lambdasweep} presents the test accuracy for varying values of $\lambda$. 

A key observation from our experiment is that as the weight attributed to the SCHaNe term increase, the performance progressively improves. We observe this trend across all tested datasets, and find that peak performance is consistently achieved when $\lambda=0.9$. We find that this optimal weighting of both terms leads to an average performance increase of 2.1\% and 2.7\% over only using the SCHaNe or CE term respectively. However, when the cross-entropy term is entirely removed (i.e., $\lambda=1$), there's a decline in accuracy from 87.0\% to 84.3\%. 


These observations underscore the complementary nature of the cross-entropy and SCHaNe loss function. Identifying an optimal balance between them provides superior performance, underscoring the importance of both terms in the overall loss function.


\subsection{Dissecting the Gains: Ablation Study on Labels and Negative Sample Impact}
We aim to verify that the performance enhancements resulting from our proposed loss function are not simply due to the incorporation of a contrastive learning term. Thus, we conducted an ablation study to delve deeper into this. Specifically, we evaluated three alternative loss functions: a supervised contrastive loss without the hard negative weighting (SupCon), an unsupervised contrastive loss (SimCLR), and vanilla cross-entropy (CE). Similar to our proposed objective, we adjusted the weighting of the contrastive term using the parameter $\lambda$ for these variants. We utilized the BEiT-3 backbone in all experiments and reported our findings on the CIFAR-FS dataset.

As illustrated in Table \ref{tab:ablation}, integrating a contrastive learning term in fact increases performance in the fine-tuning stage. Remarkably, the inclusion of label information in the supervised contrastive learning loss (SupCon) led to a significant increase in performance $12\%$ in contrast to the unsupervised approach (SimCLR). Furthermore, our proposed SCHaNe loss further improves the classification accuracy from $85.32\%$ to $87.00\%$, surpassing the results obtained with SupCon. These findings support our argument that a supervised contrastive loss, especially when paired with explicit hard negative sampling, can enhance the accuracy during fine-tuning of pre-trained models for subsequent tasks, particularly in the few-shot learning scenario.



\begin{table}[]
\centering
\begin{tabular}{@{}llccc@{}}
\Xhline{2\arrayrulewidth}
Model & Fine-tuning method & Label & HNS & Acc(\%) \\ \midrule
BEiT-3 & CE                 & \checkmark &   & 82.68       \\
BEiT-3 & CE + SimCLR        &   & &   73.99          \\
BEiT-3 & CE + SupCon        & \checkmark &  &   85.32          \\
BEiT-3 & CE + SCHaNe        & \checkmark & \checkmark & 87.00        \\ \Xhline{2\arrayrulewidth}
\end{tabular}
\caption{\textbf{Ablation study on CIFAR-FS.} Results are reported in the 1-shot few-shot learning setting using Top-1 accuracy. "CE" denotes cross-entropy loss, "Label" represents task labels, and "HNS" refers to hard negative sampling.}
\label{tab:ablation}
\end{table}

\subsection{Embedding Quality Analysis: SCHaNe vs. Cross-Entropy}
To validate the enhancements brought by the SCHaNe loss function, we perform a thorough evaluation focusing on the geometric characteristics of the generated representation spaces. We employ metrics such as isotropy~\cite{arora2016latent} and cosine similarity to provide a comprehensive understanding of these spaces. Historically, isotropy has served as an evaluative metric for representation quality~\cite{arora2016latent}. This is based on the premise that widely distributed representations across different classes in the embedding space facilitate better distinction between them. We hypothesize that our proposed objective could refine the geometric nature of the embedding space, thereby enhancing class distinction and improving transfer learning performance.

To elaborate, we examine BEiT-3 fine-tuned with vanilla cross-entropy (denoted as BEiT-3-CE) and compare it to the model when fine-tuned using our proposed loss function (denoted as BEiT-3-SCHaNe). Specifically, we evaluate three key facets of these models: 
\begin{itemize}
    \item \textbf{Distributions of cosine similarities between image pairs}: This assessment provides insights into how well the model differentiates between classes in the embedding space.
    \item \textbf{Visualization of the embedding space using the t-SNE algorithm~\cite{tsne}}: This visualization allows us to observe the separation or clustering of data points belonging to different classes, offering a spatial understanding of the embeddings.
    \item \textbf{Isotropy score as defined by \cite{mu2018allbutthetop}}: The isotropy score measures the distribution of data in the embedding space and serves as a metric for the quality of the produced embeddings.
\end{itemize}

\begin{figure}[h]
    \begin{subfigure}{0.45\linewidth}
        \includegraphics[width=1\linewidth]{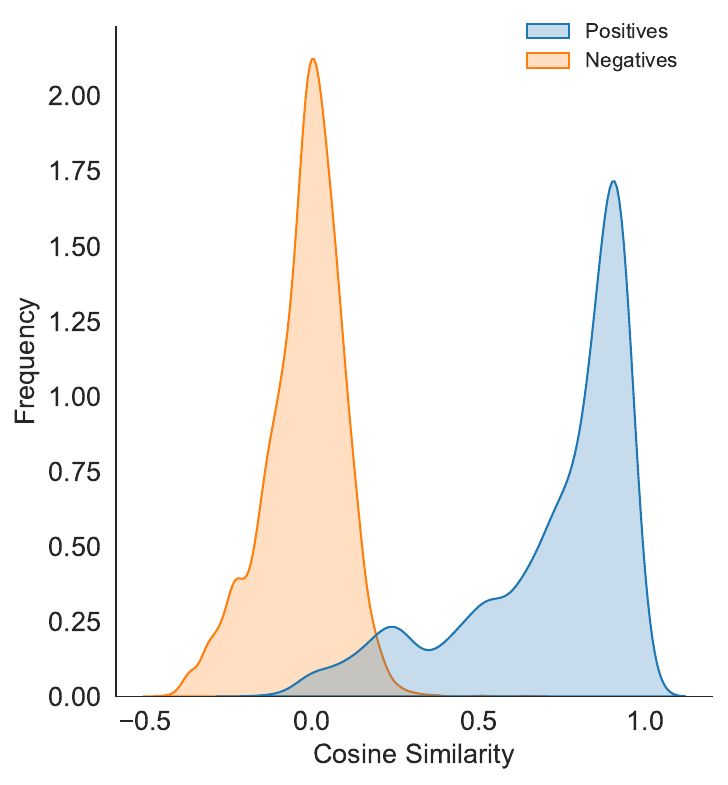}
        \subcaption[]{BEiT-3-CE}
        \label{fig:BEiT-CE_cosine}
    \end{subfigure}
\hfill 
    \begin{subfigure}{0.45\linewidth}
        \includegraphics[width=1\linewidth]{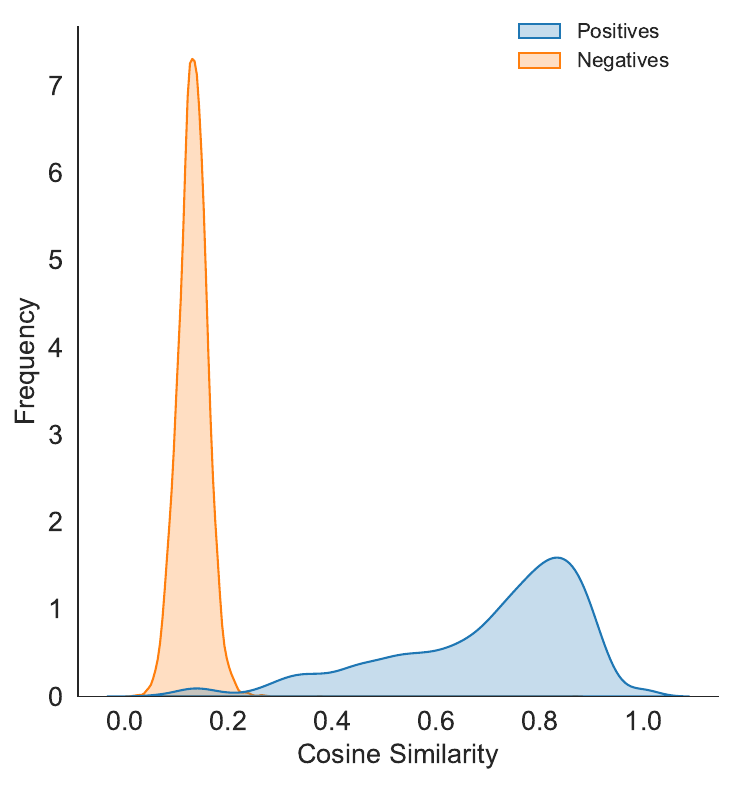}
        \subcaption[]{BEiT-3-SCHaNe}
        \label{fig:BEiT-SCHeNa_cosine}
    \end{subfigure}
\caption{\textbf{Plot of cosine similarity distribution across two random classes from CIFAR-100.} Blue represents similarities of positive samples, while orange represents similarities of negative samples.}
\label{fig:cosine}

\end{figure}

In Figure \ref{fig:cosine}, we present the pairwise cosine similarity distributions of BEiT-3-CE (Fig. \ref{fig:BEiT-CE_cosine}) and BEiT-3-SCHaNe embeddings (Fig. \ref{fig:BEiT-SCHeNa_cosine}). Specifically, we randomly select two classes from CIFAR-100 to compute cosine similarities for positive (same class) and negative pairs (different classes). Observations from these plots reveal that the BEiT-3-SCHaNe embeddings demonstrate superior separation between classes and less overlap between positive and negative samples compared to BEiT-3-CE.
\begin{figure}[h]

    \begin{subfigure}{0.48\linewidth}
        \includegraphics[width=1\linewidth]{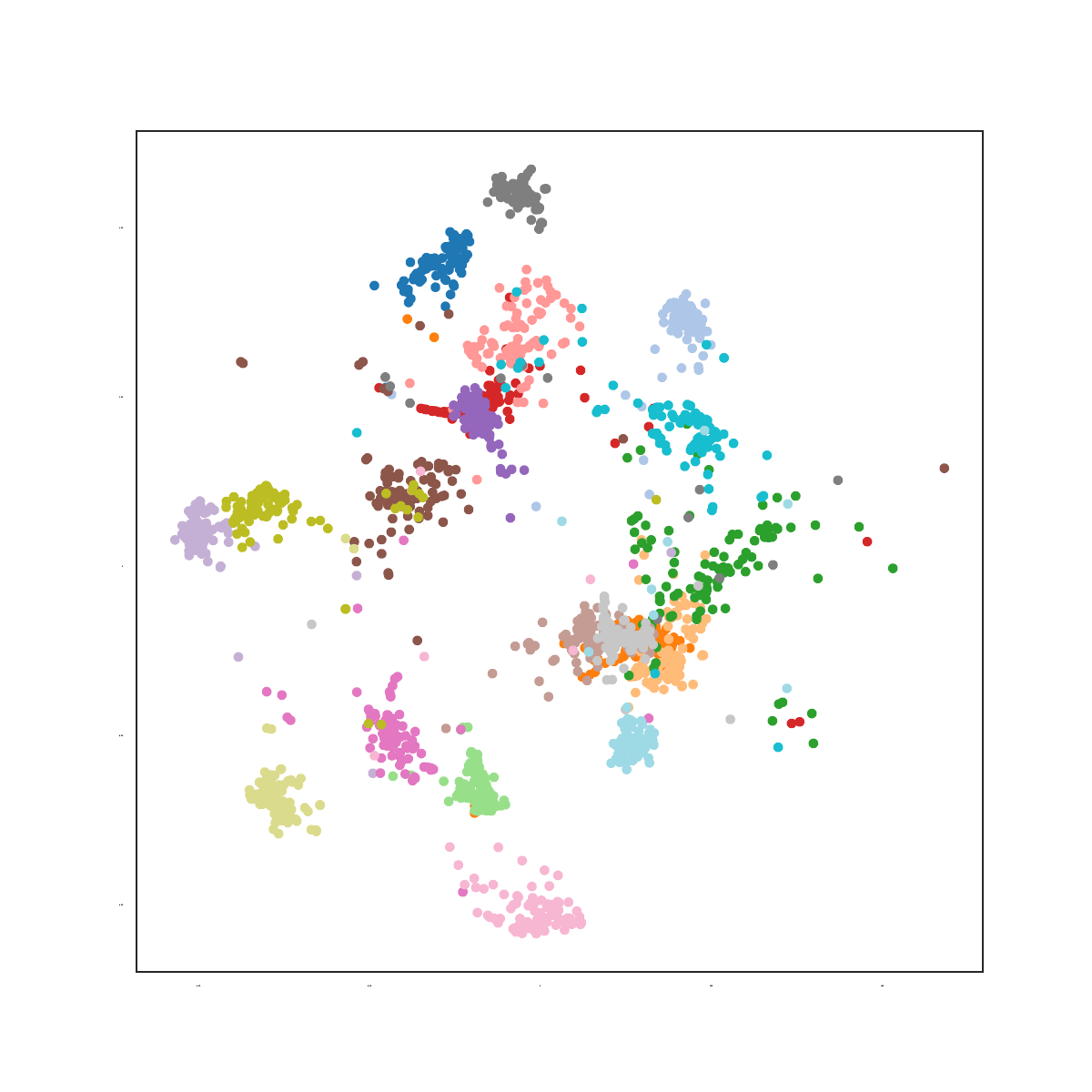}
        \subcaption[]{BEiT-3-CE}
        \label{fig:BEiT-CE_tsne}
    \end{subfigure}
\hfill 
    \begin{subfigure}{0.48\linewidth}
        \includegraphics[width=1\linewidth]{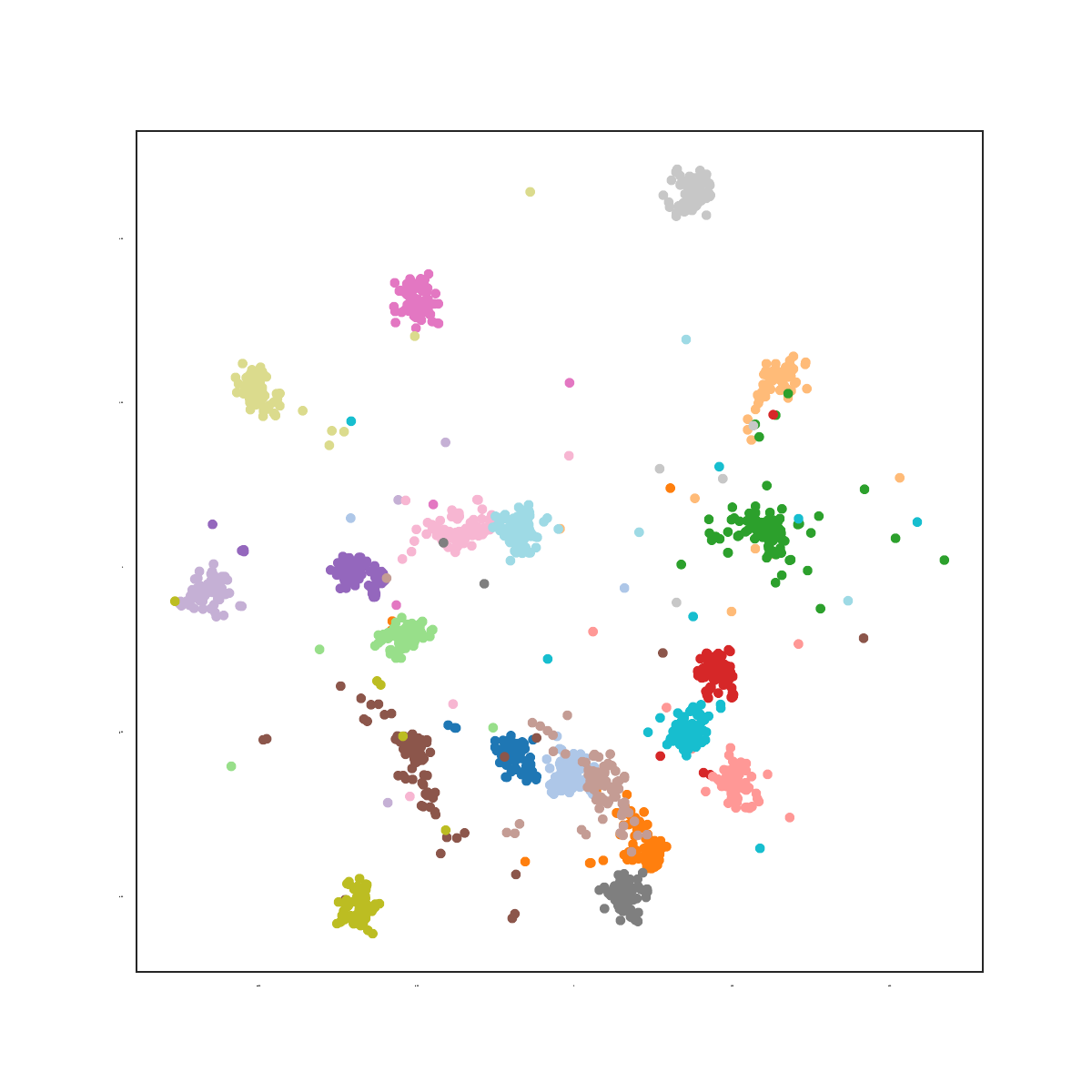}
        \subcaption[]{BEiT-3-SCHaNe}
        \label{fig:BEiT-SCHaNe_tsne}
    \end{subfigure}
\caption{\textbf{Embedding Space Visualization for BEiT-CE vs. BEiT-SCHaNe}. Displayed over twenty CIFAR-100 validation set classes using t-SNE. Each dot represents a sample, with distinct colors indicating different label classes.}
\label{fig:tsne}

\end{figure}

In Figure \ref{fig:tsne}, the t-SNE algorithm visualizes the embedding space of BEiT-3-CE and BEiT-3-SCHaNe for twenty CIFAR-100 classes. The BEiT-3-CE embeddings (Fig. \ref{fig:BEiT-CE_tsne}) display instances where the same class nodes are closely packed, but also reveal several outliers. This suggests a reduced discriminative capability. On the contrary, BEiT-3-SCHaNe embeddings (Fig. \ref{fig:BEiT-SCHaNe_tsne}) display more segregated and compact class clusters, suggesting improved classification capabilities.

Finally, we calculate the quantitative isotropy score (IS)~\cite{mu2018allbutthetop}, which is defined as follows:

\begin{equation*}
IS(\mathcal{V}) = \frac{max_{c\subset C} \sum_{v\subset V} \exp{(C^T V)}}{min_{c\subset C} \sum_{v\subset V} \exp{(C^T V)}},
    \label{eq:is}
\end{equation*}

where $V$ is a set of vectors, $C$ is the set of all possible unit vectors (i.e. any $c$ so that $||c||$ = 1) in the embedding space. In practice, $C$ is approximated by the eigenvector set of $V^TV$ ($V$ are the stacked embeddings of $v$). The larger the IS value, the more isotropic an embedding space is (i.e., a perfectly isotropic space obtains an IS score of 1).

\begin{table}[t]
\centering
\resizebox{0.47\textwidth}{!}{\begin{tabular}{@{}lccc@{}}
\Xhline{2\arrayrulewidth}
Model        & \multicolumn{1}{l}{iNaturalist2017} & \multicolumn{1}{l}{Imagenet-1k} & \multicolumn{1}{l}{Places365} \\ \midrule
BEiT3-CE     & 0.3230                              & 0.2723                          & 0.3404                        \\
BEiT3-SCHaNe & 0.9827                              & 0.9201                          & 0.9338                        \\ \Xhline{2\arrayrulewidth}
\end{tabular}}
\caption{\textbf{Comparison of Isotropy score over three datasets for BEiT-3-CE and BEiT-3-SCHaNe.} The higher value is better. A higher isotropy score means better isotropy and generalisability}
\label{tab:IS}
\end{table}

Table \ref{tab:IS} demonstrates that the IS score for BEiT-3-SCHaNe is superior to BEiT-3-CE, confirming that SCHaNe produces a more isotropic semantic space. The BEiT-3-CE embeddings are more anisotropic, implying that BEiT-3-SCHaNe embeddings more distinctly separate the different classes.

In summary, our analysis demonstrates that BEiT-3-SCHaNe produces superior embeddings compared to BEiT-3-CE. This demonstrates the limitations of using cross-entropy loss and the value-add from integrating the supervised contrastive loss function with a hard negative sampling technique (SCHaNe in this case) into the fine-tuning stage. The proposed SCHaNe contrastive loss function restructures the embedding space to enhance class distinction, addressing the limitations of the vanilla cross-entropy loss. This enhancement is particularly effective for few-shot learning scenarios, where limited labelled data requires the model to rely more on high-quality, discriminative representations. 

\section{Conclusion}
In this work, we addressed the shortcomings of cross-entropy and existing contrastive learning losses during fine-tuning. Our novel proposed objective function, the supervised contrastive loss function SCHaNe, is distinguished by its unique incorporation of hard negative sampling. By effectively leveraging labels to discern true positives from negatives and emphasizing those hard negative samples based on dissimilarity to positive counterparts, SCHaNe encourages models to generate more discerning embeddings. Our experimental results demonstrate consistent accuracy improvements across all tested datasets upon a state-of-the-art baseline, BEiT-3, both in few-shot learning and full-dataset fine-tuning settings. Importantly, SCHaNe established a new state-of-the-art for base models on ImageNet-1k with an accuracy of 86.14\%. This achievement confirms that our proposed objective effectively improves the performance of pretrained models without requiring specialized architectures or additional resources. To encapsulate, our exhaustive exploration and empirical evidence not only justify our methodological choices but also argue that SCHaNe represents a paradigm shift in enhancing the performance of pretrained models in the image classification domain. Future research will focus on applying SCHaNe in the pre-training phase and extending its applicability to other types of models such as graph neural networks and convolutional neural networks.

\bibliography{aaai23.bib}

\begin{thebibliography}{55}
\providecommand{\natexlab}[1]{#1}

\bibitem[{Arora et~al.(2016)Arora, Li, Liang, Ma, and
  Risteski}]{arora2016latent}
Arora, S.; Li, Y.; Liang, Y.; Ma, T.; and Risteski, A. 2016.
\newblock A latent variable model approach to pmi-based word embeddings.
\newblock \emph{Transactions of the Association for Computational Linguistics},
  4: 385--399.

\bibitem[{Becker and Hinton(1992)}]{becker1992self}
Becker, S.; and Hinton, G.~E. 1992.
\newblock Self-organizing neural network that discovers surfaces in random-dot
  stereograms.
\newblock \emph{Nature}, 355(6356): 161--163.

\bibitem[{Bertinetto et~al.(2019)Bertinetto, Henriques, Torr, and
  Vedaldi}]{DBLP:conf/iclr/cifarfs}
Bertinetto, L.; Henriques, J.~F.; Torr, P. H.~S.; and Vedaldi, A. 2019.
\newblock Meta-learning with differentiable closed-form solvers.
\newblock In \emph{7th International Conference on Learning Representations,
  {ICLR} 2019, New Orleans, LA, USA, May 6-9, 2019}. OpenReview.net.

\bibitem[{Cao et~al.(2019)Cao, Wei, Gaidon, Ar{\'{e}}chiga, and
  Ma}]{DBLP:journals/corr/abs-1906-07413}
Cao, K.; Wei, C.; Gaidon, A.; Ar{\'{e}}chiga, N.; and Ma, T. 2019.
\newblock Learning Imbalanced Datasets with Label-Distribution-Aware Margin
  Loss.
\newblock \emph{CoRR}, abs/1906.07413.

\bibitem[{Chen et~al.(2020)Chen, Kornblith, Norouzi, and Hinton}]{RN89}
Chen, T.; Kornblith, S.; Norouzi, M.; and Hinton, G. 2020.
\newblock A simple framework for contrastive learning of visual
  representations.
\newblock In \emph{International conference on machine learning}, 1597--1607.
  PMLR.
\newblock ISBN 2640-3498.

\bibitem[{Chuang et~al.(2020)Chuang, Robinson, Lin, Torralba, and
  Jegelka}]{DBLP:journals/corr/abs-2007-00224}
Chuang, C.; Robinson, J.; Lin, Y.; Torralba, A.; and Jegelka, S. 2020.
\newblock Debiased Contrastive Learning.
\newblock \emph{CoRR}, abs/2007.00224.

\bibitem[{Cubuk et~al.(2019)Cubuk, Zoph, Man{\'{e}}, Vasudevan, and
  Le}]{DBLP:conf/cvpr/CubukZMVL19}
Cubuk, E.~D.; Zoph, B.; Man{\'{e}}, D.; Vasudevan, V.; and Le, Q.~V. 2019.
\newblock AutoAugment: Learning Augmentation Strategies From Data.
\newblock In \emph{{IEEE} Conference on Computer Vision and Pattern
  Recognition, {CVPR} 2019, Long Beach, CA, USA, June 16-20, 2019}, 113--123.
  Computer Vision Foundation / {IEEE}.

\bibitem[{Deng et~al.(2009)Deng, Dong, Socher, Li, Li, and
  Fei{-}Fei}]{DBLP:conf/cvpr/imagenet}
Deng, J.; Dong, W.; Socher, R.; Li, L.; Li, K.; and Fei{-}Fei, L. 2009.
\newblock ImageNet: {A} large-scale hierarchical image database.
\newblock In \emph{2009 {IEEE} Computer Society Conference on Computer Vision
  and Pattern Recognition {(CVPR} 2009), 20-25 June 2009, Miami, Florida,
  {USA}}, 248--255. {IEEE} Computer Society.

\bibitem[{Devlin et~al.(2019)Devlin, Chang, Lee, and Toutanova}]{RN53}
Devlin, J.; Chang, M.; Lee, K.; and Toutanova, K. 2019.
\newblock BERT: Pre-training of Deep Bidirectional Transformers for Language
  Understanding.
\newblock In \emph{Proceedings of the 2019 Conference of the North American
  Chapter of the Association for Computational Linguistics: Human Language
  Technologies, NAACL-HLT}, 4171--4186.

\bibitem[{Dhillon et~al.(2020)Dhillon, Chaudhari, Ravichandran, and
  Soatto}]{DBLP:conf/iclr/DhillonCRS20}
Dhillon, G.~S.; Chaudhari, P.; Ravichandran, A.; and Soatto, S. 2020.
\newblock A Baseline for Few-Shot Image Classification.
\newblock In \emph{8th International Conference on Learning Representations,
  {ICLR} 2020, Addis Ababa, Ethiopia, April 26-30, 2020}. OpenReview.net.

\bibitem[{Dosovitskiy et~al.(2020)Dosovitskiy, Beyer, Kolesnikov, Weissenborn,
  Zhai, Unterthiner, Dehghani, Minderer, Heigold, and Gelly}]{RN83}
Dosovitskiy, A.; Beyer, L.; Kolesnikov, A.; Weissenborn, D.; Zhai, X.;
  Unterthiner, T.; Dehghani, M.; Minderer, M.; Heigold, G.; and Gelly, S. 2020.
\newblock An image is worth 16x16 words: Transformers for image recognition at
  scale.
\newblock In \emph{9th International Conference on Learning Representations,
  {ICLR} 2021, Virtual Event, Austria, May 3-7, 2021}.

\bibitem[{Elsayed et~al.(2018)Elsayed, Krishnan, Mobahi, Regan, and
  Bengio}]{RN97}
Elsayed, G.; Krishnan, D.; Mobahi, H.; Regan, K.; and Bengio, S. 2018.
\newblock Large margin deep networks for classification.
\newblock \emph{Advances in neural information processing systems}, 31.

\bibitem[{Ge et~al.(2018)Ge, Huang, Dong, and
  Scott}]{DBLP:journals/corr/abs-1810-06951}
Ge, W.; Huang, W.; Dong, D.; and Scott, M.~R. 2018.
\newblock Deep Metric Learning with Hierarchical Triplet Loss.
\newblock \emph{CoRR}, abs/1810.06951.

\bibitem[{Griffin, Holub, and Perona(2007)}]{griffin2007caltech}
Griffin, G.; Holub, A.; and Perona, P. 2007.
\newblock Caltech-256 object category dataset.

\bibitem[{Gunel et~al.(2020)Gunel, Du, Conneau, and Stoyanov}]{suconlan}
Gunel, B.; Du, J.; Conneau, A.; and Stoyanov, V. 2020.
\newblock Supervised Contrastive Learning for Pre-trained Language Model
  Fine-tuning.
\newblock \emph{CoRR}, abs/2011.01403.

\bibitem[{He et~al.(2021)He, Chen, Xie, Li, Dollár, and Girshick}]{RN82}
He, K.; Chen, X.; Xie, S.; Li, Y.; Dollár, P.; and Girshick, R. 2021.
\newblock Masked autoencoders are scalable vision learners.
\newblock \emph{arXiv preprint arXiv:2111.06377}.

\bibitem[{He et~al.(2019)He, Fan, Wu, Xie, and
  Girshick}]{DBLP:journals/corr/abs-1911-05722}
He, K.; Fan, H.; Wu, Y.; Xie, S.; and Girshick, R.~B. 2019.
\newblock Momentum Contrast for Unsupervised Visual Representation Learning.
\newblock \emph{CoRR}, abs/1911.05722.

\bibitem[{Hinton, Vinyals, and Dean(2015)}]{DBLP:journals/corr/HintonVD15}
Hinton, G.~E.; Vinyals, O.; and Dean, J. 2015.
\newblock Distilling the Knowledge in a Neural Network.
\newblock \emph{CoRR}, abs/1503.02531.

\bibitem[{Horn et~al.(2018)Horn, Aodha, Song, Cui, Sun, Shepard, Adam, Perona,
  and Belongie}]{DBLP:conf/cvpr/inat2017}
Horn, G.~V.; Aodha, O.~M.; Song, Y.; Cui, Y.; Sun, C.; Shepard, A.; Adam, H.;
  Perona, P.; and Belongie, S.~J. 2018.
\newblock The INaturalist Species Classification and Detection Dataset.
\newblock In \emph{2018 {IEEE} Conference on Computer Vision and Pattern
  Recognition, {CVPR} 2018, Salt Lake City, UT, USA, June 18-22, 2018},
  8769--8778. Computer Vision Foundation / {IEEE} Computer Society.

\bibitem[{Kalantidis et~al.(2020)Kalantidis, Sariyildiz, Pion, Weinzaepfel, and
  Larlus}]{DBLP:journals/corr/abs-2010-01028}
Kalantidis, Y.; Sariyildiz, M.~B.; Pion, N.; Weinzaepfel, P.; and Larlus, D.
  2020.
\newblock Hard Negative Mixing for Contrastive Learning.
\newblock \emph{CoRR}, abs/2010.01028.

\bibitem[{Khosla et~al.(2020)Khosla, Teterwak, Wang, Sarna, Tian, Isola,
  Maschinot, Liu, and Krishnan}]{RN81}
Khosla, P.; Teterwak, P.; Wang, C.; Sarna, A.; Tian, Y.; Isola, P.; Maschinot,
  A.; Liu, C.; and Krishnan, D. 2020.
\newblock Supervised contrastive learning.
\newblock \emph{Advances in Neural Information Processing Systems}, 33:
  18661--18673.

\bibitem[{Kolesnikov, Zhai, and Beyer(2019)}]{DBLP:conf/cvpr/KolesnikovZB19}
Kolesnikov, A.; Zhai, X.; and Beyer, L. 2019.
\newblock Revisiting Self-Supervised Visual Representation Learning.
\newblock In \emph{{IEEE} Conference on Computer Vision and Pattern
  Recognition, {CVPR} 2019, Long Beach, CA, USA, June 16-20, 2019}, 1920--1929.
  Computer Vision Foundation / {IEEE}.

\bibitem[{Krizhevsky, Hinton et~al.(2009)}]{krizhevsky2009learning}
Krizhevsky, A.; Hinton, G.; et~al. 2009.
\newblock Learning multiple layers of features from tiny images.

\bibitem[{Kumar et~al.(2017)Kumar, Harwood, Carneiro, Reid, and
  Drummond}]{DBLP:journals/corr/KumarHC0D17}
Kumar, B. G.~V.; Harwood, B.; Carneiro, G.; Reid, I.~D.; and Drummond, T. 2017.
\newblock Smart Mining for Deep Metric Learning.
\newblock \emph{CoRR}, abs/1704.01285.

\bibitem[{Lee et~al.(2019)Lee, Maji, Ravichandran, and
  Soatto}]{DBLP:conf/cvpr/LeeMRS19}
Lee, K.; Maji, S.; Ravichandran, A.; and Soatto, S. 2019.
\newblock Meta-Learning With Differentiable Convex Optimization.
\newblock In \emph{{IEEE} Conference on Computer Vision and Pattern
  Recognition, {CVPR} 2019, Long Beach, CA, USA, June 16-20, 2019},
  10657--10665. Computer Vision Foundation / {IEEE}.

\bibitem[{Liu et~al.(2016)Liu, Wen, Yu, and Yang}]{DBLP:conf/icml/LiuWYY16}
Liu, W.; Wen, Y.; Yu, Z.; and Yang, M. 2016.
\newblock Large-Margin Softmax Loss for Convolutional Neural Networks.
\newblock In Balcan, M.; and Weinberger, K.~Q., eds., \emph{Proceedings of the
  33nd International Conference on Machine Learning, {ICML} 2016, New York
  City, NY, USA, June 19-24, 2016}, volume~48 of \emph{{JMLR} Workshop and
  Conference Proceedings}, 507--516. JMLR.org.

\bibitem[{Mosbach, Andriushchenko, and
  Klakow(2020)}]{DBLP:journals/corr/abs-2006-04884}
Mosbach, M.; Andriushchenko, M.; and Klakow, D. 2020.
\newblock On the Stability of Fine-tuning {BERT:} Misconceptions, Explanations,
  and Strong Baselines.
\newblock \emph{CoRR}, abs/2006.04884.

\bibitem[{Mu and Viswanath(2018)}]{mu2018allbutthetop}
Mu, J.; and Viswanath, P. 2018.
\newblock All-but-the-Top: Simple and Effective Postprocessing for Word
  Representations.
\newblock In \emph{International Conference on Learning Representations}.

\bibitem[{Nar et~al.(2019{\natexlab{a}})Nar, Ocal, Sastry, and
  Ramchandran}]{DBLP:journals/corr/abs-1901-08360}
Nar, K.; Ocal, O.; Sastry, S.~S.; and Ramchandran, K. 2019{\natexlab{a}}.
\newblock Cross-Entropy Loss and Low-Rank Features Have Responsibility for
  Adversarial Examples.
\newblock \emph{CoRR}, abs/1901.08360.

\bibitem[{Nar et~al.(2019{\natexlab{b}})Nar, Ocal, Sastry, and
  Ramchandran}]{RN98}
Nar, K.; Ocal, O.; Sastry, S.~S.; and Ramchandran, K. 2019{\natexlab{b}}.
\newblock Cross-entropy loss and low-rank features have responsibility for
  adversarial examples.
\newblock \emph{arXiv preprint arXiv:1901.08360}.

\bibitem[{Nilsback and Zisserman(2008)}]{DBLP:conf/icvgip/flowers}
Nilsback, M.; and Zisserman, A. 2008.
\newblock Automated Flower Classification over a Large Number of Classes.
\newblock In \emph{Sixth Indian Conference on Computer Vision, Graphics {\&}
  Image Processing, {ICVGIP} 2008, Bhubaneswar, India, 16-19 December 2008},
  722--729. {IEEE} Computer Society.

\bibitem[{Oreshkin, L{\'{o}}pez, and Lacoste(2018)}]{DBLP:conf/nips/fc100}
Oreshkin, B.~N.; L{\'{o}}pez, P.~R.; and Lacoste, A. 2018.
\newblock {TADAM:} Task dependent adaptive metric for improved few-shot
  learning.
\newblock In Bengio, S.; Wallach, H.~M.; Larochelle, H.; Grauman, K.;
  Cesa{-}Bianchi, N.; and Garnett, R., eds., \emph{Advances in Neural
  Information Processing Systems 31: Annual Conference on Neural Information
  Processing Systems 2018, NeurIPS 2018, December 3-8, 2018, Montr{\'{e}}al,
  Canada}, 719--729.

\bibitem[{Parkhi et~al.(2012)Parkhi, Vedaldi, Zisserman, and
  Jawahar}]{DBLP:conf/cvpr/pets}
Parkhi, O.~M.; Vedaldi, A.; Zisserman, A.; and Jawahar, C.~V. 2012.
\newblock Cats and dogs.
\newblock In \emph{2012 {IEEE} Conference on Computer Vision and Pattern
  Recognition, Providence, RI, USA, June 16-21, 2012}, 3498--3505. {IEEE}
  Computer Society.

\bibitem[{Ravi and Larochelle(2017)}]{DBLP:conf/iclr/RaviL17}
Ravi, S.; and Larochelle, H. 2017.
\newblock Optimization as a Model for Few-Shot Learning.
\newblock In \emph{5th International Conference on Learning Representations,
  {ICLR} 2017, Toulon, France, April 24-26, 2017, Conference Track
  Proceedings}. OpenReview.net.

\bibitem[{Ren et~al.(2018)Ren, Triantafillou, Ravi, Snell, Swersky, Tenenbaum,
  Larochelle, and Zemel}]{DBLP:conf/iclr/tieredimagenet}
Ren, M.; Triantafillou, E.; Ravi, S.; Snell, J.; Swersky, K.; Tenenbaum, J.~B.;
  Larochelle, H.; and Zemel, R.~S. 2018.
\newblock Meta-Learning for Semi-Supervised Few-Shot Classification.
\newblock In \emph{6th International Conference on Learning Representations,
  {ICLR} 2018, Vancouver, BC, Canada, April 30 - May 3, 2018, Conference Track
  Proceedings}. OpenReview.net.

\bibitem[{Robinson et~al.(2021)Robinson, Chuang, Sra, and
  Jegelka}]{DBLP:conf/iclr/RobinsonCSJ21}
Robinson, J.~D.; Chuang, C.; Sra, S.; and Jegelka, S. 2021.
\newblock Contrastive Learning with Hard Negative Samples.
\newblock In \emph{9th International Conference on Learning Representations,
  {ICLR} 2021, Virtual Event, Austria, May 3-7, 2021}. OpenReview.net.

\bibitem[{Russakovsky et~al.(2015)Russakovsky, Deng, Su, Krause, Satheesh, Ma,
  Huang, Karpathy, Khosla, and Bernstein}]{RN91}
Russakovsky, O.; Deng, J.; Su, H.; Krause, J.; Satheesh, S.; Ma, S.; Huang, Z.;
  Karpathy, A.; Khosla, A.; and Bernstein, M. 2015.
\newblock Imagenet large scale visual recognition challenge.
\newblock \emph{International journal of computer vision}, 115(3): 211--252.

\bibitem[{Schroff, Kalenichenko, and
  Philbin(2015)}]{DBLP:journals/corr/SchroffKP15}
Schroff, F.; Kalenichenko, D.; and Philbin, J. 2015.
\newblock FaceNet: {A} Unified Embedding for Face Recognition and Clustering.
\newblock \emph{CoRR}, abs/1503.03832.

\bibitem[{Snell, Swersky, and Zemel(2017)}]{DBLP:conf/nips/SnellSZ17}
Snell, J.; Swersky, K.; and Zemel, R.~S. 2017.
\newblock Prototypical Networks for Few-shot Learning.
\newblock In Guyon, I.; von Luxburg, U.; Bengio, S.; Wallach, H.~M.; Fergus,
  R.; Vishwanathan, S. V.~N.; and Garnett, R., eds., \emph{Advances in Neural
  Information Processing Systems 30: Annual Conference on Neural Information
  Processing Systems 2017, December 4-9, 2017, Long Beach, CA, {USA}},
  4077--4087.

\bibitem[{Sohn(2016)}]{npairloss}
Sohn, K. 2016.
\newblock Improved Deep Metric Learning with Multi-class N-pair Loss Objective.
\newblock In Lee, D.; Sugiyama, M.; Luxburg, U.; Guyon, I.; and Garnett, R.,
  eds., \emph{Advances in Neural Information Processing Systems}, volume~29.
  Curran Associates, Inc.

\bibitem[{Song et~al.(2015)Song, Xiang, Jegelka, and
  Savarese}]{DBLP:journals/corr/SongXJS15}
Song, H.~O.; Xiang, Y.; Jegelka, S.; and Savarese, S. 2015.
\newblock Deep Metric Learning via Lifted Structured Feature Embedding.
\newblock \emph{CoRR}, abs/1511.06452.

\bibitem[{Suh et~al.(2019)Suh, Han, Kim, and Lee}]{suh2019stochastic}
Suh, Y.; Han, B.; Kim, W.; and Lee, K.~M. 2019.
\newblock Stochastic class-based hard example mining for deep metric learning.
\newblock In \emph{Proceedings of the IEEE/CVF Conference on Computer Vision
  and Pattern Recognition}, 7251--7259.

\bibitem[{Szegedy et~al.(2015)Szegedy, Vanhoucke, Ioffe, Shlens, and
  Wojna}]{DBLP:journals/corr/SzegedyVISW15}
Szegedy, C.; Vanhoucke, V.; Ioffe, S.; Shlens, J.; and Wojna, Z. 2015.
\newblock Rethinking the Inception Architecture for Computer Vision.
\newblock \emph{CoRR}, abs/1512.00567.

\bibitem[{Tian, Krishnan, and Isola(2020)}]{DBLP:conf/eccv/TianKI20}
Tian, Y.; Krishnan, D.; and Isola, P. 2020.
\newblock Contrastive Multiview Coding.
\newblock In Vedaldi, A.; Bischof, H.; Brox, T.; and Frahm, J., eds.,
  \emph{Computer Vision - {ECCV} 2020 - 16th European Conference, Glasgow, UK,
  August 23-28, 2020, Proceedings, Part {XI}}, volume 12356 of \emph{Lecture
  Notes in Computer Science}, 776--794. Springer.

\bibitem[{Tian et~al.(2020)Tian, Wang, Krishnan, Tenenbaum, and
  Isola}]{DBLP:conf/eccv/TianWKTI20}
Tian, Y.; Wang, Y.; Krishnan, D.; Tenenbaum, J.~B.; and Isola, P. 2020.
\newblock Rethinking Few-Shot Image Classification: {A} Good Embedding is All
  You Need?
\newblock In Vedaldi, A.; Bischof, H.; Brox, T.; and Frahm, J., eds.,
  \emph{Computer Vision - {ECCV} 2020 - 16th European Conference, Glasgow, UK,
  August 23-28, 2020, Proceedings, Part {XIV}}, volume 12359 of \emph{Lecture
  Notes in Computer Science}, 266--282. Springer.

\bibitem[{Van~der Maaten and Hinton(2008)}]{tsne}
Van~der Maaten, L.; and Hinton, G. 2008.
\newblock Visualizing data using t-SNE.
\newblock \emph{Journal of machine learning research}, 9(11).

\bibitem[{Vinyals et~al.(2016)Vinyals, Blundell, Lillicrap, Kavukcuoglu, and
  Wierstra}]{DBLP:conf/nips/miniimagenet}
Vinyals, O.; Blundell, C.; Lillicrap, T.; Kavukcuoglu, K.; and Wierstra, D.
  2016.
\newblock Matching Networks for One Shot Learning.
\newblock In Lee, D.~D.; Sugiyama, M.; von Luxburg, U.; Guyon, I.; and Garnett,
  R., eds., \emph{Advances in Neural Information Processing Systems 29: Annual
  Conference on Neural Information Processing Systems 2016, December 5-10,
  2016, Barcelona, Spain}, 3630--3638.

\bibitem[{Wah et~al.(2011)Wah, Branson, Welinder, Perona, and
  Belongie}]{WahCUB_200_2011}
Wah, C.; Branson, S.; Welinder, P.; Perona, P.; and Belongie, S. 2011.
\newblock {The Caltech-UCSD Birds-200-2011 Dataset}.
\newblock Technical Report CNS-TR-2011-001, California Institute of Technology.

\bibitem[{Wang et~al.(2022)Wang, Bao, Dong, Bjorck, Peng, Liu, Aggarwal,
  Mohammed, Singhal, Som, and Wei}]{beit3}
Wang, W.; Bao, H.; Dong, L.; Bjorck, J.; Peng, Z.; Liu, Q.; Aggarwal, K.;
  Mohammed, O.~K.; Singhal, S.; Som, S.; and Wei, F. 2022.
\newblock Image as a Foreign Language: BEiT Pretraining for All Vision and
  Vision-Language Tasks.
\newblock \emph{CoRR}, abs/2208.10442.

\bibitem[{Wu et~al.(2017)Wu, Manmatha, Smola, and
  Kr{\"{a}}henb{\"{u}}hl}]{DBLP:journals/corr/WuMSK17}
Wu, C.; Manmatha, R.; Smola, A.~J.; and Kr{\"{a}}henb{\"{u}}hl, P. 2017.
\newblock Sampling Matters in Deep Embedding Learning.
\newblock \emph{CoRR}, abs/1706.07567.

\bibitem[{Yalniz et~al.(2019)Yalniz, J{\'{e}}gou, Chen, Paluri, and
  Mahajan}]{DBLP:journals/corr/abs-1905-00546}
Yalniz, I.~Z.; J{\'{e}}gou, H.; Chen, K.; Paluri, M.; and Mahajan, D. 2019.
\newblock Billion-scale semi-supervised learning for image classification.
\newblock \emph{CoRR}, abs/1905.00546.

\bibitem[{Yun et~al.(2019)Yun, Han, Oh, Chun, Choe, and
  Yoo}]{DBLP:journals/corr/abs-1905-04899}
Yun, S.; Han, D.; Oh, S.~J.; Chun, S.; Choe, J.; and Yoo, Y. 2019.
\newblock CutMix: Regularization Strategy to Train Strong Classifiers with
  Localizable Features.
\newblock \emph{CoRR}, abs/1905.04899.

\bibitem[{Zhang et~al.(2017)Zhang, Ciss{\'{e}}, Dauphin, and
  Lopez{-}Paz}]{DBLP:journals/corr/abs-1710-09412}
Zhang, H.; Ciss{\'{e}}, M.; Dauphin, Y.~N.; and Lopez{-}Paz, D. 2017.
\newblock mixup: Beyond Empirical Risk Minimization.
\newblock \emph{CoRR}, abs/1710.09412.

\bibitem[{Zhang et~al.(2020)Zhang, Wu, Katiyar, Weinberger, and
  Artzi}]{DBLP:journals/corr/abs-2006-05987}
Zhang, T.; Wu, F.; Katiyar, A.; Weinberger, K.~Q.; and Artzi, Y. 2020.
\newblock Revisiting Few-sample {BERT} Fine-tuning.
\newblock \emph{CoRR}, abs/2006.05987.

\bibitem[{Zhou et~al.(2014)Zhou, Lapedriza, Xiao, Torralba, and
  Oliva}]{DBLP:conf/nips/places}
Zhou, B.; Lapedriza, {\`{A}}.; Xiao, J.; Torralba, A.; and Oliva, A. 2014.
\newblock Learning Deep Features for Scene Recognition using Places Database.
\newblock In Ghahramani, Z.; Welling, M.; Cortes, C.; Lawrence, N.~D.; and
  Weinberger, K.~Q., eds., \emph{Advances in Neural Information Processing
  Systems 27: Annual Conference on Neural Information Processing Systems 2014,
  December 8-13 2014, Montreal, Quebec, Canada}, 487--495.

\end{thebibliography}
\end{document}